\documentclass[a4paper,10pt,5p]{elsarticle}
\journal{Expert Systems with Applications}

\usepackage{float}
\restylefloat{figure}
\restylefloat{table}

\usepackage{amsfonts}
\usepackage{setspace}
\usepackage{array}
\usepackage{amssymb}
\usepackage{graphicx}
\usepackage{url}
\usepackage{multicol}
\usepackage{stfloats}
\usepackage{subfigure}
\usepackage[usenames]{color}
\usepackage{flushend}
\usepackage{booktabs}
\usepackage{changes}
\usepackage{eurosym}
\usepackage{rotating}
\usepackage{color}
\usepackage{graphicx}
\usepackage{cite}
\usepackage{lscape}
\usepackage{tablefootnote}
\usepackage{tabularx,ragged2e}
\graphicspath{ {Figures/} }
\usepackage{hyperref}

\usepackage{cleveref}
\crefname{section}{§}{§§}
\Crefname{section}{§}{§§}
\usepackage{lipsum}
\makeatletter
\def\ps@pprintTitle{%
 \let\@oddhead\@empty
 \let\@evenhead\@empty
 \def\@oddfoot{}%
 \let\@evenfoot\@oddfoot}
\makeatother
\begin{document}
\begin{frontmatter}

\title{Transfer Learning for Multi-lingual Tasks - a Survey}

\author[label1]{Amir Reza Jafari \corref{cor1}}
\ead{amirjafari.teh@ut.ac.ir}

\author[label1]{Behnam Heidary \corref{cor1}}
\ead{bheidary@ut.ac.ir}

\author[label2]{Reza Farahbakhsh}
\ead{reza.farahbakhsh@it-sudparis.eu}

\author[label1,label3]{Mostafa Salehi \corref{cor2}}
\ead{mostafa\_salehi@ut.ac.ir}

\author[label4]{Mahdi Jalili}
\ead{mahdi.jalili@rmit.edu.au}

\cortext[cor1]{Equal contribution}
\cortext[cor2]{Corresponding author}
\address[label1]{New Sciences and Technologies, University of Tehran, Tehran, Iran}
\address[label2]{Institut Polytechnique de Paris, Telecom SudParis, Evry, France}
\address[label3]{Institute for Research in Fundamental Science (IPM), Tehran, Iran}
\address[label4]{School of Engineering, RMIT University, Australia}

\begin{abstract}

These days different platforms such as social media provide their clients from different backgrounds and languages the possibility to connect and exchange information. It is not surprising anymore to see comments from different languages in posts published by international celebrities or data providers. In this era, understanding cross languages 
content and multilingualism in natural language processing (NLP) are hot topics, and multiple efforts have tried to leverage existing technologies in NLP to tackle this challenging research problem. In this survey, we provide a comprehensive overview of the existing literature with a focus on transfer learning techniques in multilingual tasks. We also identify potential opportunities for further research in this domain.  
\end{abstract}
\begin{keyword}
Transfer Learning, BERT, NLP, NLU, Multilingual task, low resource languages, language models.
\end{keyword}
\end{frontmatter}

\section{Introduction}
\label{sec:Introduction}

The phenomenon of multilingualism in different NLP tasks is one of the most exciting and demanding topics in this area. During the past decade, these topics have been the center of attention in the linguistic and computer science community alongside the increasing use of transfer learning in NLP.
This task is significantly more critical with the extensive usage of social media and massive engagement of end-users across the world to trending topics. As multilingualism is gaining massive attention in order to reach better performance in NLP multilingual tasks and applications, an overview about the history of transfer learning in language models, published main language models, and more specifically, multilingual, cross-lingual, and even language-specific models is necessary to step in this field.

\begin{figure}[ht]
    \centering
	\includegraphics[width=9cm,height=7.9cm]{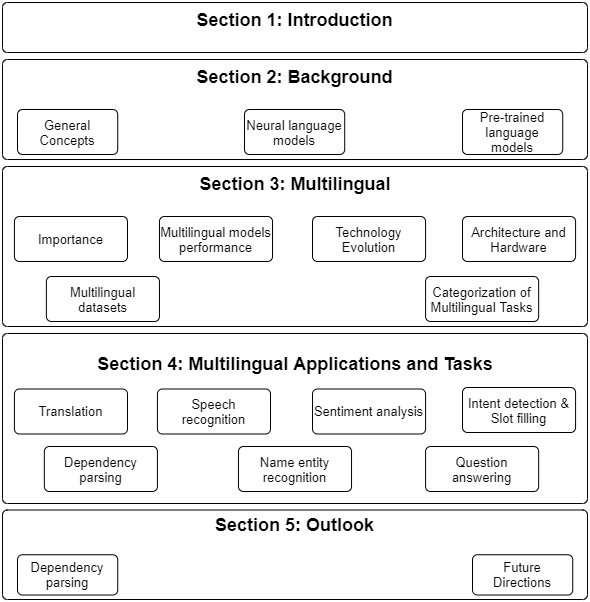}
	\vspace{-0.5cm}
	\caption{content of the survey and components of each section }
	\label{survey_struc} 
\end{figure}

In our survey, we mainly focused on multilingualism models and tasks; see figure \ref{survey_struc} as an illustration of the main components presented in this paper. First, we started by reviewing the main concepts and a brief history of language models. We then categorized them into three main groups in Section \ref{sec:Background}. Later in Section \ref{Multilingual}, we focused more on reviewing architecture and structure from a multilingual perspective and specified the importance of cross-lingual and multilingual models in NLP. Furthermore, we introduced available datasets for each application to help those who want to work on a specific domain in this subject. Since evaluating these language models is possible with analyzing different pre-defined NLP applications, we focused on these applications in section \ref{Applications} and reviewed the existing literature that evaluates their models on these applications in different languages. In Section \ref{sec:challenges}, we provide future directions and the challenges that exist in this subject, and we tried to provide a good perspective for future studies.

The concept of transfer learning and its solutions in machine learning and data mining has been reviewed in \citep{Weiss2016}, and more detailed history about the evolution of transfer learning in NLP can be found in \citep{ruder-etal-2019-transfer}. \citep{Qiu220} provides a comprehensive survey of pre-trained models in NLP and categorizes the existing pre-trained models based on taxonomy. Also, \citep{Dabre2019} \citep{Pikuliak2020} reviewed multilingualism in specific NLP tasks such as machine translation and text processing. Our survey aims to review a more general overview of multilingualism in different tasks and introduce those language models that deal with different languages or a language with lower resources. Table \ref{surveys_compare} shows our main focus in this work and how it compares with other surveys. We will not provide many details about transfer learning. Our survey can be helpful for those with sufficient knowledge about transfer learning who are interested in applying it to multilingual models and tasks.

\begin{table*}
	\scriptsize
	\begin{tabular}{|m{4cm}|m{6cm}|m{5cm}|m{1.2cm}|}\hline
		\textbf{Title} & \textbf{Main Focus of the study} & \textbf{How differentiate it with our paper} & \textbf{Published}
		\\ \hline
		A survey of transfer learning \citep{Weiss2016} & 
		Mainly focused on transfer learning paradigm and its current solutions and applications applied to transfer learning & An overview of transfer learning with less details and more focus on this paradigm in multilingual models and applications & 2016
		\\ \hline
	    A survey of cross-lingual word embedding models \citep{Ruder2017} &
	    
	    provide a comprehensive typology of cross-lingual word embedding models and compare their data requirements and objective functions & we focused on outputs of models and only talk about structure and word embedding enough to help readers to understands outputs & 2017
		\\ \hline
		Transfer Learning in Natural Language Processing (NLP) \citep{ruder-etal-2019-transfer} & This survey represents an effort at providing a succinct yet complete understanding of the recent advances in NLP using deep learning in with a special focus on detailing transfer learning and its potential advantages & We assumed that readers have a basic knowledge of transfer learning and we focused on transfer learning in multilingual models & 2019
		\\ \hline
		A survey of multilingual neural machine translation \citep{Dabre2019} & This survey present an in-depth survey of existing literature on MNMT and also categorize various approaches based on the resource scenarios as well as underlying modeling principles & We have a more general overview of multilingual tasks which include machine translation too but not limited to a specific task & 2019
		\\ \hline
		Cross-lingual learning for text processing: A survey \citep{Pikuliak2021} & a comprehensive table of all the surveyed papers with various data related to the cross-lingual learning techniques they use & we have a model perspective and focused on multilingual language models more & 2021
		\\ \hline
	\end{tabular}
	\caption{Recent surveys in field of Transfer learning and multilingual NLP tasks}
	\label{surveys_compare}

\end{table*}

\section{Background}
\label{sec:Background}
\newcolumntype{N}{>{\centering\arraybackslash}m{0.1\linewidth}}
\newcolumntype{L}{>{\centering\arraybackslash}m{0.6\linewidth}}
\newcolumntype{F}{>{\centering\arraybackslash}m{0.3\linewidth}}
\newcolumntype{T}{>{\centering\arraybackslash}m{0.2\linewidth}}

By applications of transfer learning on language models, a new era has emerged in the NLP domain. Most performance analysis techniques on NLP-related tasks focus on languages with sufficiently large available data, and those with low resources are often kept out of attention. Before reviewing the history of language models and introducing transfer learning in this subject, it is necessary to overview some basic concepts. In this section, we provide a brief overview of general concepts, including language models and transfer learning.

\subsection{General Concepts}

\subsubsection{Language Models}
Language Modeling (LM) is one of the main parts of NLP tasks, which uses various probabilistic techniques for predicting a word or a sequence of words in a sentence. The importance of LM in NLP is undeniable, especially in terms of multilingual models which contain many NLP tasks, such as machine translation\citep{Vaswani2013}, question answering\citep{Bouziane2015}, speech recognition\citep{Mikolov2010} or sentiment analysis \citep{Mantyla2018}. From the statistical point of view, LM is a learning process to predict the probability distribution of a sequence of words occurring in a sentence. In fact, LM works with analyzing the text in data by learning the features and characteristics of a language with suitable algorithms and then understanding phrases and predict the next words in sentences by probabilistic analysis\citep{Osborne2019} \citep{Smith2017}.

\subsubsection{Transfer Learning}
Transfer Learning is one of the Machine learning approaches, where the information gained from pre-training a model with general tasks is reused in other related tasks for improving efficiency and faster fine-tuning\citep{ruder-etal-2019-transfer}. This approach was introduced as a machine learning approach with the introduction of ImageNet in 2010\citep{JiaDeng2009} as a successful large CNN model. With fine-tuning deep neural networks, more than 14 million images have been divided into more than 20,000 categories. Transfer Learning has been used in large number of studies in different NLP Applications, and propsed state-of-the-art result in that case like sentiment analysis 

\subsubsection{Multilingual and Cross-Lingual Analysis}
These two terms, which are usually used interchangeably in most works, can be defined as follow:

Multilingual/Cross-lingual learning is a part of transfer learning that focuses on transferring knowledge from one language with usually higher available resources to another language with lower resources. This concept may lead to better performance in many downstream tasks, especially in languages lacking valuable data. In general, We can look at these concepts from two perspectives:

1: Multilingual usually deals with models. We define this concept as a model pre-trained on different language datasets that check performance on related downstream tasks.
Cross-lingual usually comes with learning a model based on a high resource language and then use and evaluate this model for low-resource language for different NLP tasks\citep{Pikuliak2021}.

2: In terms of cross-lingual embedding, the same vector projection is used for similar words in different languages as a semantic view.  In Multilingual embeddings, just using the same embeddings for different languages is considered without assurance of interaction between different languages. In addition, In cross-lingual, we have a query in one language, and the aim is to retrieve the document in another language. However, in Multilingual, in addition to this, the focus is on the models that deal with multiple languages.

\subsubsection{Zero-shot Learning}
Zero-shot Learning (ZSL) is a type of classification problem where a classifier is trained on a specific set of labels in different classes during training, and then it evaluates the samples that have not been previously observed \citep{Larochelle2008}. 
ZSL in multilingual tasks refers to classifying data based on few or even no labeled examples in under-resourced languages with training on multiple languages with noticeable resources.   

To narrow down the ZSL in NLP downstream tasks, ZSL plays an important role specifically in the field of cross-lingual. For example, in \citep{Pushp2018} ZSL is used for text classification for generalizing models on new unseen classes after training to learn the relationship between a sentence and the embedding of its tags. ZSL is also used for news sentiment classification by assigning the sentiment category to news in other languages without any training data required after training on Slovene news \citep{Pelicon2020}. \citep{Ma2020} and \citep{Banerjee2020} used ZLS for the Question-answering task to generalize it to unseen questions. Since intent-detection plays a crucial role in question-answering, \citep{Xia2018} studied the zero-shot intent detection problem to detect user intents for no labeled utterances data. For the task of Entity recognition with portable domain, \citep{Guerini2018} presents a zero-shot learning approach for entity recognition of users' talks which not annotated during training, and for Dependency parsing, \citep{Tran2019} analyses the ZSL approach for Multilingual Sentence Representations

\begin{figure*}[ht]
	\centering
	\includegraphics[width=\textwidth,height=3.5cm]{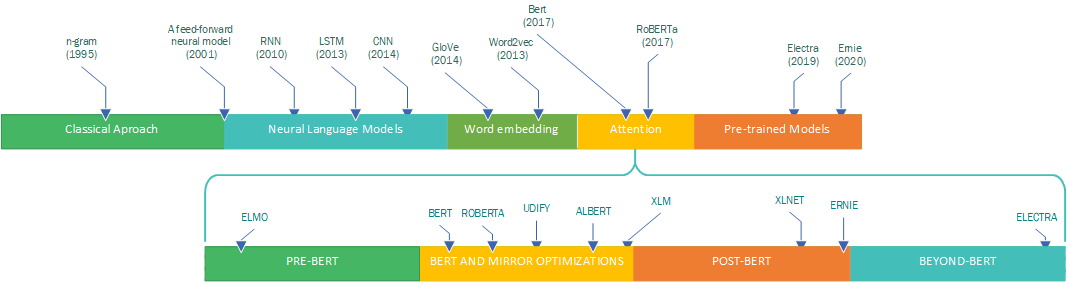}
	\caption{Evolution of Transfer Models in NLP}
	\label{model_evolution}
\end{figure*}

\subsection{Neural language models}
{The early methods used in NLP research were mainly based on probabilistic language models, such as n-gram \citep{Banchs2006}. These simple language models aim to predict the next word in a sequence by assigning a probability to a sequence of words.}

{The first binding of neural networks with language modeling was proposed in 2001 \citep{Bengio2001}. With simultaneously learning of distributed representation and probability function for each word, this model improves n-gram models and also can use longer context as an input.}

{The influence of the main types of neural networks in NLP started with the introduction of Recurrent neural networks (RNNs) \citep{Mikolov2010}. In RNNs, the output of the previous step is used as an input for predicting the next word. This method has shown remarkable performance in solving such problems using hidden layers. Since RNNs are difficult to train, Long short term memory (LSTM) \citep{Hochreiter1997} has become more popular for language modeling\citep{Graves2013}}

More recently, convolutional neural networks (CNNs) have also been used in NLP research. \citep{Kalchbrenner2014} proposed a Dynamic k-Max Pooling network over linear sequences to extract sentences' feature graph for semantic modeling of sentences. The main advantages of these networks are supporting varying length sentences as input and also their applicability to any language. For sentence-level classification tasks, \citep{Kim2011} used CNN with little hyperparameter tuning and static vectors to improve performance in NLP tasks, such as sentiment analysis and question classification. By using dilation in the convolutional layers to increase its receptive field, the ByteNet proposed a mechanism to address the source's variant lengths and the target in context \citep{Kalchbrenner2016}. The combination of both CNNs and LSTMs is used for dimensional sentiment analysis \citep{Wang2016}, and for faster train and test time, QRNNs was proposed over LSTM \citep{Bradbury2017} 

Since many machine learning algorithms in text processing are incapable of processing strings or plain text in their raw format, they need to convert these inputs into numerical vectors. Assigning words with the same semantic and syntactic view to the same vectors as a basic definition of word embeddings, a distributed representation of words \citep{Bengio2003b} learned the joint probability function of sequences of words and trained them in a neural language model. In 2013 \citep{Mikolov2013} introduced one of the most popular techniques (called word2vec) to learn word embeddings using neural network methods such as Skip Gram and Common Bag Of Words (CBOW). Word2vec uses both algorithms to learn weights that act as word vector representations used for different NLP tasks.

Another word embedding method for vector representation of words, called GloVe, is based on unsupervised learning algorithm\citep{Brennan2017}. GloVe uses a different mechanism and equations to create the embedding matrix. Instead of training on the entire sparse matrix or individual context windows, this model tries to train on word-word co-occurrence probabilities in a large corpus. As a result, this model outperforms on word analogy task and similarity tasks and named entity recognition because of the vector space structure that has been introduced \citep{Brennan2017}

\subsection{Pre-trained Language Models}
{Collbert and Weston \citep{Collobert2008} proposed a single convolutional neural network architecture that can be referred as a starting point in pre-trained models. The output of this neural network architecture for a given sentence can be used for NLP downstream tasks.}

{With introduction of transfer learning, a revolution in Language models architecture has begun, leading to significant improvements in downstream NLP tasks' performance. Bidirectional training of transformers, which was the BERT model's innovation, enabled training on a text sequence to be either from left-to-right or combined left-to-right and right-to-left. In the transformer mechanism, the input text is first processed by an encoder, and then the decoder predicts the task's goal. Therefore, the encoder reads the entire input sequence at once, allowing the model to learn the context from all previous and next tokens. This often provides high accuracy.} 

{The reputation of transfer learning had a significant influence on pre-trained models. It made building NLP models easier since transfer learning made it possible to train a model on one dataset at first and then performing different NLP tasks on a different dataset. This phenomenon is becoming more popular, especially in multilingualism, since the structure that is needed for this is adopted well with the transfer learning concept.}

\begin{table*}
	\scriptsize
	\begin{tabular}{|m{3cm}|m{2cm}|m{1.5cm}|m{1.1cm}|m{8cm}|}\hline
		\textbf{Model} & \textbf{Type} & \textbf{Focused Language} & \textbf{Publish Year} & \textbf{Input Corpus}
		\\ \hline
		BERT \citep{Devlin2018} & Base model & English & 2018 &16GB of uncompressed text:BookCorpus (800M words)English Wikipedia (2500M words)
		\\ \hline
		RoBERTa \citep{Liu2019a} & Base model & English & 2019 &160GB text:BookCorpus (800M words) (16GB)CC-News (63M English news articles) (76GB)OpenWebText (Web content extracted from URLs shared on Reddit) (38GB)Stories (subset of CommonCrawl data) (31GB)
		\\ \hline
		ELECTRA \citep{Manning2020} & Base model & English & 2020 &For experiments (Same Data as BERT) : 3.3 billion tokens from WIkipedia and BooksCorpusFor Language model: extend the BERT dataset to 33B tokens by including data form: ClueWeb; CommonCrawl; Gigaword
		\\ \hline
		ERNIE \citep{Zhang2019} & Base model & English & 2020 &Processed Wikipedia Eng (4; 500M subwords and 140M entities)
		\\ \hline
		AlBERT \citep{Of2020} & Base model & English & 2020 &16GB of uncompressed text consist of:BookCorpus (800M words)English Wikipedia (2500M words)
		\\ \hline
		UDify \citep{Kondratyuk2019a}& Base model & multilingual & 2019 & full Universal Dependencies v2.3 corpus available on LINDAT, Arabic NYUAD, English ESL,French FTB, Hindi English HEINCS, Japanese BC-CWJ
		\\ \hline
		XLNet \citep{Yang2019b} & Base model & English & 2019 & RACE Dataset, SQuAD, GLUE Dataset, ClueWeb09-B Dataset
		Following
		\\ \hline
		mBERT & Multilingual Models & Cross-lingual & 2018 &  Wikipedia, MultiUN, IIT Bombay corpus, OPUS, EUbookshop, OpenSubtitles, GlobalVoices, Kytea and PyThaiNLP5
		\\ \hline
		XLM \citep{Lample2019}& Multilingual Models & Cross-lingual & 2019 &  Wikipedia, MultiUN, IIT Bombay corpus, OPUS, EUbookshop, OpenSubtitles, GlobalVoices, Kytea and PyThaiNLP5
		\\ \hline
		XLM-R \citep{Conneau}& Multilingual Models & Cross-lingual & 2019 &  2.5TB of text from CommonCrawl
		\\ \hline
		CamemBERT \citep{Martin2019} &Language-Specific model& French & 2019 & 138GB of uncompressed text and 32.7B SentencePiece tokens consis of:French text extracted from CommonCrawlUnshuffled version of the French OSCAR corpus
		\\ \hline
		RobBERT \citep{Delobelle2020} & Language-Specific model & German  & 2020 & 39GB of uncompressed text consis of:Dutch Section of OSCAR corpus (6.6B words) (39GB of texts)  
		\\ \hline
		FlauBERT \citep{Le2019} &Language-Specific model & French & 2019 & 71GB of cleaned Unicode-Normalized:Common Crawl
		\\ \hline
		AraBERT \citep{Antoun2020} & Language-Specific model & Arabic& 2020 & 24GB of text (70M sentences) cosist of:  1- Arabic Corpus: 1.5 billion words     2- OSIAN: 3.5 billion articles  
		\\ \hline
		BERTje \citep{DeVries2019} & Language-Specific model & Dutch  & 2019 &  Books: a collection of novels (4.4GB) TwNC a Dutch News Corpus (2.4GB) SoNaR-500 a reference corpus (2.2GB)4 Dutch news websites (1.6GB) Wikipedia dump (1.5GB)Total: 12 GB; 2.4B token
		\\ \hline
		ALBERTo \citep{Polignano2019}&Language-Specific model &Italian & 2019 & TWITA:from twitter's official streaming API; 200M tweets and 191GB raw data
		\\ \hline
		PhoBERT \citep{Nguyen2020}&Language-Specific model& Vietnamese &2020& 20GB texts: Vietnamese Wikipedia corpus (1GB)-(19GB) is a subset of a Vietnamese news corpus
		\\ \hline
		BERT for Finnish \citep{Virtanen2019} & Language-Specific model& Finnish   & 2019  &  Yle corpus, an archive of news and STT corpus of newswire articles
		\\ \hline
	\end{tabular}
	\caption{Main Characteristics of the existing models.}
	\label{lang_models}
\end{table*}

{We categorized the existing pre-trained language models into the three main groups:}
\begin{itemize}
	\item \textit{Base Models:} Those types of language model which introduce a new structure in LM
	\item \textit{Multilingual Models:} Those types of language model which deal with multiple languages
	\item \textit{Language-specific Models:} Those types of language model which focus on a specific languages rather than English
\end{itemize}

\subsubsection{Base Models}
The term `Base Model' refers to the models that gained a huge amount of attention by introducing a new structure or changing the previous architecture. We focused more on Bert and post-Bert models that are shown in figure \ref{model_evolution}. 
In 2018, Google's AI language team introduced a Bidirectional Encoder Representations from Transformers called "BERT" which was a revolution in the field of pre-trained models \citep{Devlin2018}. This pre-training model contains learning from unlabeled text jointly in both the left and right directions. This innovation led to outstanding improvement in a wide range of NLP tasks. 
A year after BERT, Facebook introduced a new optimized method called "RoBERTa" based on the masking strategy used in Bert and changed a number of key parameters in that model. In addition, increasing dataset size and training time turned out to be a critical improvement in results. Also, another key change in this model in comparison with BERT was removing "Next Sentence Prediction" which was marked as an unnecessary task by RoBERTa\citep{Liu2019a}. 
Another model that we can consider as a base model in our presented model types is "ERNIE" which is Enhanced representation through knowledge integration that it outperforms Google's BERT in multiple language tasks concentrating in Chinese language\citep{Zhang2019}.

\subsubsection{Multilingual Models}
{Since many language models focused on a single language representation, multilingual models gained attention in this field. After successfully proposing BERT by Google, the multilingual version of BERT was published a year after. This model, called "mBERT", supported sentence representation for 104 languages. This model outperformed previous works in many multilingual tasks. Concentrating on semantic aspect of mBERT, \citep{Libovicky2019} shows that splitting mBERT representation into two separate components: language-specific and language-neutral. The second component has high accuracy in less difficult tasks such as ord-alignment and sentence retrieval.}
{Another model based on Transformers with masked language modeling (MLM) objective like BERT, is XLM which trained with translation Language Modeling to learn different languages similar representations \citep{Lample2019}.
XLM structure is based on BERT, but as roBERTa \citep{Liu2019a} proved improvement in results with changing parameters compared to BERT, a new multilingual model called XLM-R had been published, which eliminated translation Language Modeling task in XLM and instead, trained roBERTa on a bigger multilingual dataset containing 100 languages \citep{Conneau}}

\subsubsection{Language-Specific Models}
Although multilingual models showed high performance in multilingual tasks, researchers showed that concentrating on a specific language and fine-tuning for particular tasks in that language can lead to better results in sub-tasks. For example, CamemBERT model, which is a French pre-trained model based on roBERTa, showed that learning on French data and fine-tuning only for French outperform other multilingual models like mBERT and UDify\citep{Martin2019}.
In Table \ref{lang_models} some other language specific models has been represented which shows that proposing a language-specific model for each particular language is a new trend in this field.

\section{Multilingual}
\label{Multilingual}

Increasing the efficiency of transformers and technology shifts in processing units has made it possible to provide language models that can advance multiple languages. In this section, we discuss the importance of multilingual models and analyze their maturity. We propose a big picture of these models side by side with the monolingual models. We also investigate research studies from two viewpoints, historical and model characteristics. We also review the architecture, performance, hardware requirements, and language features of these models.

\subsection{Importance of the multilingual tasks}
Practical NLP applications are often developed for English language because it is impossible to train large and precise language models in languages with a small labeled dataset. The importance of modeling for such languages in unexpected situations 
has been investigated. However, language models in low-resource languages are not limited to emergencies, and for providing a wide range of new NLP-dependent technology services, which are now primarily done in the context of deep neural networks, language models are necessary.

Cross-lingual models use large unlabeled datasets of one language to build a language model that can be fine-tuned in another language with a small corpus, to perform much better in the target language.

\subsection{Multilingual Models Performance} \label{mlperformamce}
In this part, we review the studies that have examined the capabilities of multilingual models. Some focused on the strengths of these models and applications that have good performance; other ones showed NLP tasks in which the performance of multilingual models was inferior to monolingual models. Table \ref{tab:mm_perf} compared these studies.

\begin{table*}
	\scriptsize
	\begin{tabular}{|m{3cm}|m{1cm}|m{3cm}|m{3cm}|m{3.5cm}|m{2cm}|}
		\hline
		\textbf{Title of the Study} & 
		\textbf{Model} &
		\textbf{Dataset} &
		\textbf{Evaluation Criteria} &
		\textbf{Results} &
		\textbf{Ref.}  
		\\\hline
		    How multilingual is Multilingual BERT? &
		    mBert &
		    104 languages Wikipedia &
		    examines the multilingual capability &
		    mBert has an amazing performance in cross-lingual  tasks &
		     \citep{Pires2019}
		\\\hline
		How Language-Neutral is Multilingual BERT? &
		mBert & 
		use a pre-trained mBERT and train on specific language Wikipedia, WMT14 &
		semantic properties of mBERT &
		mBert representations split into a language-specific  and a language-neutral component that each one are suitable for specific tasks &
		\citep{Libovicky2019}
		\\\hline
		    Beto, Bentz, Becas: The surprising cross-lingual effectiveness of Bert&
		    mBert &
		    Reuters corpus covering 8 languages &
		    evaluate as a zero-shot cross-lingual model on multiple languages and NLP tasks & 
		    fine-tuned hyper parameters mBert has an amazing performance &
		    \citep{Wu2020}
		\\\hline
		    Is Multilingual BERT Fluent in Language Generation?&
		    mBert &
		    Universal Dependencies treebanks &
		    ability to substitute monolingual models &
		    inefficiency of multilingual models in text generation task &
		     \citep{Ronnqvist2019}
		\\\hline
		Cross-lingual ability of multilingual BERT: An empirical study &
		B-BERT &
		XNLI and LORELEI &
		cross-lingual ability covering linguistic properties and similarities of languages, model architecture and inputs and training objectives&
		B-BERT amazing results in cross-lingual applications &
		\citep{Karthikeyan2019}
		\\\hline
	\end{tabular}
	\centering
	\caption{Studies focused on cross-lingual aspects of multilingual models}
	\label{tab:mm_perf}
\end{table*}

Pires et al. in \citep{Pires2019} examines the multilingual capabilities of the mBert model, and for this purpose, the model was pre-trained on the Wikipedia data set collected from more than 100 languages. The model was then fine-tuned with task-specific supervised data in one language and tested for performance in another. The results of this study show that mBert has incredible performance in cross-lingual tasks. Factors influencing the model's performance mentioned lexical overlap and typological similarity. However, the model performed well in languages with different scripts.
Another study deals with the semantic features of the mBert model and divides the resulting model into two parts related to specific language and general language. The second part performed well in tasks such as word alignment and exact sentence retrieval but is not suitable for machine translation applications \citep{Libovicky2019}. 
\\
\citep{Wu2020} evaluates mBERT as a zero-shot cross-lingual model on about 40 different languages and five different NLP tasks: natural language inference, document classification, NER, part-of-speech tagging, and dependency parsing. They show that with fine-tuned hyperparameters, mBert has an excellent performance in the mentioned tasks.\\
On the other hand, some studies have shown the inefficiency of multilingual models in some applications \citep{Ronnqvist2019}. 
 \\
Some research studies have shown BERT's great performance in cross-lingual applications, which is even more surprising because of the absence of a cross-lingual objective during the training phase. \citep{Karthikeyan2019} examines the effect of different components of the BERT model on its cross-lingual performance in three different aspects: linguistic features (lexical and structural), model architecture, and the format of the inputs and training objectives. They showed that the lexical similarity (similar words or similar parts in words) of languages has a negligible effect on performance, and instead, the depth of the network is much more effective. Another conclusion is that the depth of the network and the total number of the parameters in the architectural features of the model were more critical than the Multi-head layers of attention. They also showed that NSP and tokenizing at the character or word level reduce model performance in cross-lingual.\\
Some studies work on specific task optimization of multilingual models. CLBT model \citep{Wang2020} focus on dependency parsing that point to lexical properties. 
 
\subsection{Technology Evolution}
The process of technology development in the field of multilingual models can be studied from a historical perspective (evolution in time) or a model perspective, which will be detailed in this section. 

\subsubsection{Historical Review}
\begin{figure*}[ht]
	\centering
	\includegraphics[width=\textwidth]{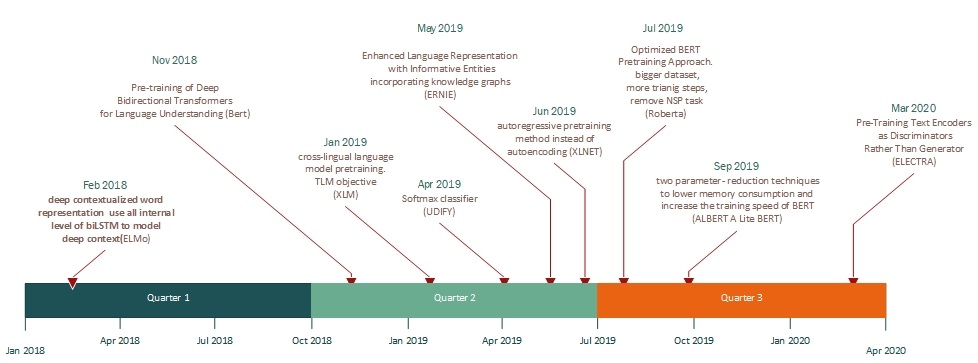}
	\caption{Evolution of Linguistic Technologies (from Time Perspective)}
	\label{fml_tl_t_p}
\end{figure*}

In terms of historical evolution in time, multilingual models have gone through a challenging process (Figure 3). At first, models like ELMO \citep{Peters2018} remained loyal to bidirectional LSTM  and performed well. Then, transformers \citep{Vaswani2017} were introduced and led the architecture and performance of the models for some time. Transformers increased the need for parallel execution and performance in various tests by replacing the recursive architecture with an attention mechanism and increasing the requirement for processing resources and training time. The BERT model was released and improved later in many other works. For example, the ALBERT model uses methods to reduce the number of parameters to provide the heavy BERT model in lighter versions.
Shortly, researchers took a separate path in terms of design and architecture then proposed fresh models such as XLNET using autoregressive model or ELECTRA that pre-train text encoder as the generator.

\subsubsection{Model Perspective}
\begin{figure*}[ht]
	\centering
	\includegraphics[width=\textwidth]{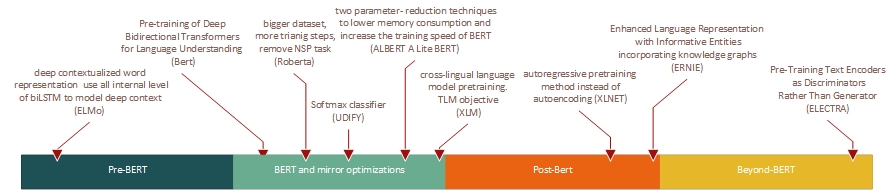}
	\caption{Evolution of Linguistic Technologies (from Modeling Perspective)}
	\label{fig_multilingu_timeline_mdl_pres}
\end{figure*}

From the model point of view, according to Figure \ref{fig_multilingu_timeline_mdl_pres}, multilingual research studies fall into four categories:
The first generation that came before introducing BERT, such as ELMO, shifted the results by using all the output of the Bidirectional LSTM inner layers.
In the second generation, BERT and its minor improvements are categorized, using more extensive data sets and changing pre-train tasks, classifier changes and optimizations are some of the changes seen in the Roberta\citep{Liu2019a}, UDIFY \citep{Kondratyuk2019a}, Albert\citep{Of2020}, and XLM \citep{Lample2019}.

In the Post-BERT era, models had significant modifications, including XLENT\citep{Yang2019b} and ERNIE\citep{Zhang2019}. The first uses an auto-regressive pre-train instead of an Auto-encoder.

In the next stage, we have models such as ELECTRA\citep{Manning2020} that took a relatively different path than the BERT-based models. For example, in ELECTRA, the encoder is trained as a discriminator instead of a generator.

GPT \citep{Peters2018a}  and mBert \citep{Devlin2018} focus on learning contextual word embeddings. These learned encoders are still needed to represent words in context by downstream tasks. Besides, various pre-training tasks are also proposed to learn PTMs for different purposes.

\textbf{UDify}:
This model uses over 120 Universal Dependencies \citep{11234/1-2895} treebanks in more than 70 languages and fine-tuned BERT on all datasets as a single one. That shows state-of-the-art universal POS, UFeats, Lemmas, UAS, and LAS scores. Hence can be assumed, multilingual multi-task model. \citep{Kondratyuk2019a}
 
 \textbf{mBert}:
Multilingual BERT published same time with BERT, support over 100 languages. Technically, It is just BERT trained on Wikipedia text of many languages. For the content size bias resistance for different languages, low resource languages were oversampled and general languages undersampled.
 
\textbf{XLM}:
This study was presented to evaluate Pre-trained cross-lingual models (XLMs) and suggested two methods for pre-training. The first method is unsupervised pre-training based on monolingual data, and the second method is pre-training based on multilingual data. Evaluations were performed in the XLNI \citep{Conneau2018a} and WMT'16 tasks \citep{papineni-etal-2002-bleu}. 
Another innovation of this research \citep{Lample2019} is the introduction of several objectives for pre-learning. They used MLM and Causal Language Modeling (CLM) for unsupervised learning, which examined its proper performance. They also used translation language modeling objective (TLM) alongside MLM, which is essentially an extension of MLM in the BERT model, using a set of parallel sentences instead of consecutive sentences.

\textbf{XLM-R}:
A self-supervised model uses RoBERTa objective task on a CommonCrawl dataset\footnote{https://commoncrawl.org} contains the unlabeled text of 100 languages with a token number of five times more than RoBERTa. The advantage of this model is that, unlike XLM, it does not require parallel entry, so it is scalable. \citep{Conneau}

\textbf{XLNet}: 
If we categorize unsupervised learning in two types of autoregressive and autoencoding, the XLNet model \citep{Yang2019b} focuses on autoregressive models that attempts to estimate the probability distribution of the text. In contrast, autoencoding models such as BERT try to reconstruct the original data by seeing incomplete data generated by covering some sentence tokens. The advantage of these models over autoregressive models is that they can advance the learning process in both directions. Their disadvantage is that guessed tokens are independent of uncovered tokens, which are very far from the features of natural language. The XLNet model tries to take advantage of both categories using permutations of all possible modes instead of focusing on one-way learning in AR models. In this way, it uses the content of both sides in guessing a token. Also, because it does not use incomplete data, it is unnecessary to face the difference of tokens in the pre-training and fine-tuning phase, which is the weakness of AE models.

\subsection{Architecture and Hardware Requirements}
From an architectural point of view, most models use an architecture similar to BERT-base or BERT-large \citep{Manning2020} \citep{Of2020} \citep{Yang2019b}. This group of research studies uses a combination of transformer and attention layers, in which the attention layers play a vital role in embedding the meaning and context of words.

On the other hand, there are other models that had extended the BERT architecture \citep{Zhang2019} or acted differently \citep{Peters2018} \citep{Lample2019}.

Another aspect of comparing models is batch size. From this point of view, models similar to BERT use 8,000 batch size, while other models, such as ERNIE, use size of 512.

As shown in previous research works 
Transfer learning is affiliated with computing resources. In this section, we show which models are more efficient in the manner of hardware requirements.\\
As shown in the table \ref{tbh}, several research teams have introduced basic models based on the Transformers architecture, which differ in terms of architecture, the total number of model parameters, and the hardware used.
In terms of the hardware processing units used, the models use TPUs or GPUs. Although different models use proprietary combinations of hardware, in some cases, such as the XLNet, up to 512 TPU has been used for less than three days, which according to the CEO of Hologram AI, cost 245,000\$ and produced 5 tons of CO2. To defeat BERT in 18 of 20 tasks \citep{Synced2019}.\\
Regarding the number of model parameters, we can name from the small version of the Electra model with 14 million parameters to Albert Large with 235 million parameters.

\begin{table*}
	\scriptsize
	\begin{tabular}{|m{2cm}|m{3cm}|m{6cm}|m{2cm}|m{3cm}|}
		\hline
		\textbf{Model} & \textbf{Team} & \textbf{Architecture Details} & \textbf{Parameters Number} & \textbf{Hardware} 
		\\ \hline \hline
		BERT \citep{Devlin2018} & Google AI & Based on the Transformer architecture; deeply bidirectional model Base:12 layers (transformer blocks); 12 attention headsLarge: 24 layers (transformer blocks); 16 attention heads  & Base:110M Large:335M &4 to 16 Cloud TPUs; 1 TPU; 64 gb ram
		\\ \hline
		ELECTRA \citep{Manning2020}&Stanford University; Google Brain&Transformers (Same as BERT)  -Generators and Discriminators- ELECTRA-small: 256 hidden dimensions (instead of 768) ; 128 token embedding (instead of 768); 128 sequence length (instead of 512) & Small:14M           Large:110M & Small : 1 V100 GPU             Large: 16TPUv3s 
		\\ \hline
		ERNIE \citep{Zhang2019} &
		Tsinghua University, Huawei Noah’s Ark Lab& 
		BERT + two multi-head self-attention. 6 layer textual encoder, 6 layer knowledgeable encoder, hidden dimension of token embedding=768, hidden dimension of entity embedding= 100, self-attention heads: Aw = 12, Ae = 4& 
		114M &  
		8 NVIDIA-2080Ti 
		\\ \hline
		AlBERT \citep{Of2020} &Google Research; Toyota Technological Institute at Chicago&  4 models: base with 12 layers and 768 hiddens, large with 24 layers and 1024 hiddens, xlarge with 24 layers and 2048, xxlarge with 24 layers and 4096 hiddens   & Base:12M Large:18M XL:60M     XXL:235M& 64 to 512 Cloud TPU V3 
		\\ \hline
		ELMo \citep{Peters2018} & Allen Institute; Allen School of CS; University of Washington & 2 BiLSTM layers with 4096 units and 512 dimension projections and a residual connection from the first to second layer  & 499M \citep{Li2019} & 3 GTX 1080 \citep{elmohardware}
		\\ \hline
		XLM \citep{Lample2019}& Facebook AI Research & 1024 hidden units, 8 heads, GELU activation & XLM-15:250M XLM-17:570M XLM-100:570M & 64 Volta GPUs for the language modeling tasks, and 8 GPUs for the MT tasks
		\\ \hline
		XLNet  \citep{Yang2019b}& Carnegie Mellon University; Google AI Brain Team & same as BERT-Large, batch size of 8192 & 110M & 512 TPU v3
		\\ \hline
	\end{tabular}
	\centering
	\caption{Architecture and Hardware requirements of the Base Models}
	\label{tbh} 
\end{table*}
\subsection{Multilingual Datasets}
In table \ref{dataset_table}, the available multilingual datasets are introduced, and for each dataset, in addition to a short description, we provided the evaluation metrics and the task that is used in previous studies.

\begin{table*}
	\scriptsize
	\begin{tabular}{|m{3cm}|m{6cm}|m{2cm}|m{2cm}|m{3cm}|}
		\hline
		\textbf{Dataset Name} & 
		\textbf{Description}&
        \textbf{Task}  & 
        \textbf{Languages} & 
        \textbf{Used in}
		\\ \hline \hline
		SNIPS \citep{coucke2018snips} &  
		contains several day to day user command categories (e.g. play a song, book a restaurant).&  
		Slot Filling and Intent Detection & 
		EN DE ES FR IT JA KO PT\_BR PT\_PT & \citep{Babu} \citep{Chen} \citep{Coucke} \citep{Yang2021}
		\\ \hline
		MTOP \citep{Li2020} & parallel multilingual task-oriented semantic parsing corpora. crowd-sourced  100k example in 11 domain and 117 intent used for 3 way evaluation: in-language, multilingual, zero-shot & Slot Filling and Intent Detection &  EN DE FR ES HI TH & \citep{Desai2021} \citep{Kaliamoorthi2021}
		\\ \hline
 		Multilingual ATIS \citep{8461905}& ATIS dataset subset translated to 2 languages by a human expert to show that results can surpass the proposed approaches with only a few labeled tokens. & 
 		Slot Filling and Intent Detection &  
 		EN HI TR & -
 		\\ \hline
 		Facebook’s multilingual dataset \citep{Schuster2019} &
 		under 60k annotated utterances about alarm-reminder and weather &
 		Slot Filling and Intent Detection & 
 		EN TH ES & 
 		\citep{8907842}
        \\ \hline
        CommonCrawl &
        Over petabyte crawled web data from 2008 and released it publicly &
        MLM &
        more than 40 languages &
        \citep{Conneau} \citep{8894409} \citep{9316706} \citep{9164940}
		\\ \hline
		MultiATIS++ \citep{Xu2020a} & extend train and test set of the English ATIS & 
		Slot Filling and Intent Detection &  EN ES DE FR PT HI ZH JA TR & -
		\\ \hline
		
		XED \citep{Kajava} & 
		A multilingual fine-grained emotion dataset &
		Sentiment analysis & 
		Mainly EN , Finnish and 30 additional languages & -
		\\ \hline
		WikiAnn \citep{WikiAnn} & 
		cross-lingual name tagging and linking based on Wikipedia articles.
		assigning a coarse-grained or fine-grained type to each mention, and link it to an English Knowledge Base if it is linkable & 
		NER & 295 languages & \citep{Ponti} \citep{Xue}
		\\ \hline
			CODAH  \citep{chen-etal-2019-codah}& adversarially-constructed evaluation dataset with 2.8k questions for testing common sense.
			model challenging extension to the SWAG dataset, which tests commonsense knowledge using sentence-completion questions that describe situations observed in video. & 
			Question Answering & EN & \citep{Lin} \citep{Yan} \citep{Bartolo}
		\\ \hline
		HotpotQA \citep{yang-etal-2018-hotpotqa} &  dataset with 113k Wikipedia-based question-answer pairs &
		Question Answering & EN & \citep{Beltagy} \citep{Ainslie} \citep{Joshi}
		\\ \hline
		NewsQA \citep{trischler2016newsqa} &  reading comprehension dataset of over 100K human-generated question-answer pairs from over 10K news articles from CNN, with answers consisting of spans of text from the corresponding articles &  
		Question Answering & EN & \citep{Joshi} \citep{He} \citep{Beltagy}
		\\ \hline
	\end{tabular}
	\centering
	\caption{Available multilingual dataset for different tasks}
	\label{dataset_table} 
\end{table*}

\subsection{Categorization of Multilingual Tasks}

\begin{figure}[ht]
	\centering
	\includegraphics[scale=0.35]{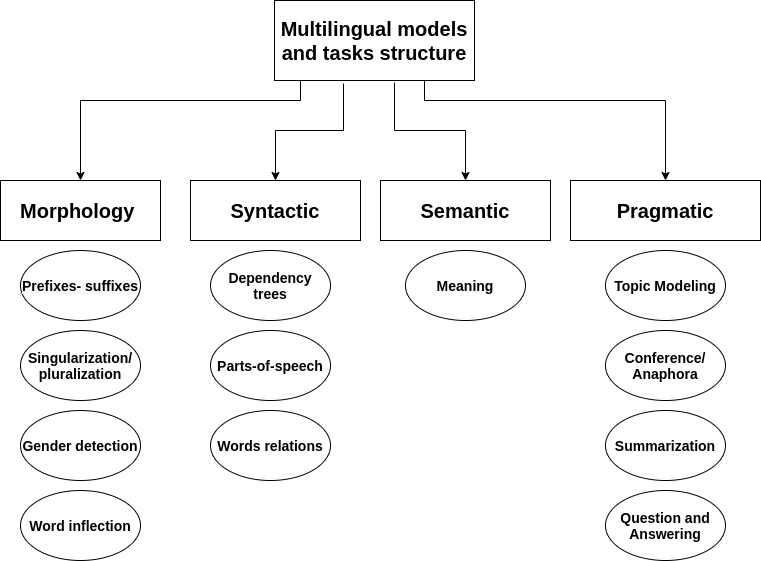}
	\caption{Categorization of Multilingual Tasks}
	\label{Multi_Categ}
\end{figure}

As shown in Figure \ref{Multi_Categ},we can categorized linguistic domain to be considered for multilingual tasks from several perspectives:

\textit{- From Morphology point of view}: Since morphology deals with the formation of words and the relation of words together, defining this category is meaningful in the multilingual task because this formation varies in different languages but can have many common properties too. The morphological structure of words usually consists of prefixes/suffixes, singularization/pluralization, gender detection, word inflection (words modification in order to express grammatical categories).

\textit{- From Syntax point of view}: Syntactic perspective in multilingual tasks refers to words relation and combination to form a bigger language unit such as sentences, clauses, and phrases. In everyday life, this view is more commonly known as the grammatical view. Alongside the relation between words, part of speech and dependency tree are considered in this category. 

\textit{- From Semantics point of view}: This view refers to the meaning of words and sentences. Semantic perspective is one of the main categories in linguistics for the multilingual task because semantic structure and relation of words and sentences are essential features of any language. 
\textit{- From Pragmatics point of view}: The pragmatic perspective in multilingual tasks deals with the contribution of context to meaning. There are several hot topics such as topic modeling, coreference-anaphora, summarization, and question-answering in NLP that are considered in this perspective.

\section{Multilingual Applications and Tasks}
\label{Applications}

With the large amount of data being generated every day in different forms, such as unstructured data, emails, chats, and tweets, the role of NLP tasks and applications gain ever-increasing importance. Analyzing data using these applications will help businesses to gain valuable insights. Trending topics like elections and Covid-19 often result in increased activity in the content generation on social media, requiring attention from the NLP community. For low-resources languages, some applications become more challenging to analyze. However, in general, NLP successfully has been used in different types of applications such as virtual assistants, speech recognition, sentiment analysis, chatbots, etc. \citep{NLPApp}. 

As an example, Google Translate, which is a free multilingual machine translation service developed by Google, is powered by NLP behind the scene. Amazon Alexa or Google Assistant uses speech recognition and NLP techniques such as question answering, text classification, and machine translation to help users achieve their goals. Even in the digital marketing industry, analyzing data using these techniques helps the community understand customers' interests and generate accurate reports based on business needs.

A key value of our study is the importance of reviewing different NLP and application tasks, not just in the English language but in other languages. Since many NLP models and applications try to cover multiple languages, having preliminary knowledge about the works on these applications and tasks from a multilingual perspective would help researchers follow their works. In this section, we analyze NLP applications and tasks from a multilingual point of view separately.

With the help of transfer learning, one may reach a good performance on many NLP tasks, not only in a high-resource language like English but many low-resources languages. Since language models other than English are getting more attention in academia and industry, new research studies concentrate more on the multilingual aspect of NLP in different tasks.
In some cases, transfer learning from a multilingual model to a language-specific model can improve performance in many downstream tasks. \citep{Kuratov2019} uses this approach for the Russian language, which resulted in improving performance on s reading comprehension, paraphrase detection, and sentiment analysis tasks. Furthermore, the training time of this model decreased compare to multilingual models.
One of the aspects of text analysis is the style of the text. Many factors, such as formality markers, emotions, and metaphors, influence the analysis of the style of the text. \citep{Kang2019} provides a benchmark corpus (xSLUE) containing text in 15 different styles and 23 classification tasks as an online platform for cross-style language understanding and evaluation. This research shows that there are many ways to develop low-resource or low-performance styles and other applications such as cross-style generation.
 
\begin{figure}[ht]
	\centering
	\includegraphics[scale=0.3]{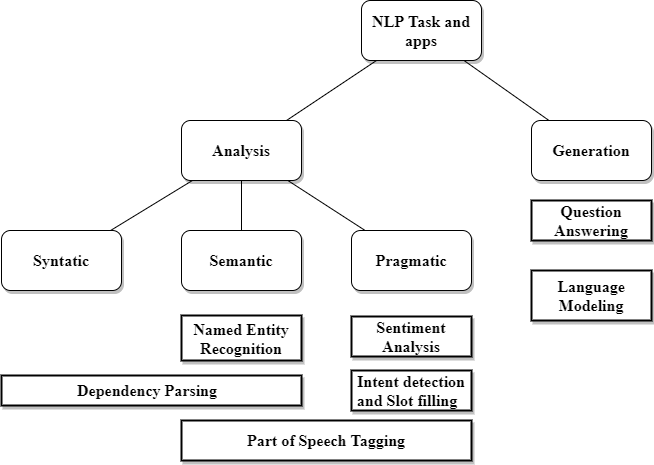}
	\caption{NLP Tasks and Application Categories}
	\label{xx}
\end{figure}

Another challenge in NLP applications, especially in low-resource languages, is detecting hate speech. \citep{Stappen2020} developed an architecture for pre-trained Transformers to examine cross-lingual zero-shot and few-shot learning. This model with the novel attention-based classification block AXEL uses transformers technique for English and Spanish datasets.

Also, \citep{Tawalbeh2020} uses transfer learning with BERT and RNN to represent shared tasks on multilingual offensive language.

\subsection{Translation}
The importance of translation in the field of NLP is undeniable. Especially in the case of multilingual, this service is in the spotlight. Most of these models train on a single model (mostly English) and try to translate to other languages. Facebook AI introduced "M2M-100" \citep{Fana}, a Many-to-Many multilingual translation model that translates directly between any pair language among a pool of 100 languages.
Using zero-shot systems, \citep{Lakew2018} explores closeness between languages focusing both on automatic standard metrics (BLEU and TER).
\subsection{Speech Recognition}
Much research has been conducted in the field of Speech Recognition, mainly focusing on deep neural networks and RNNs \citep{Huang2013} \citep{Mohan2015} \citep{Zhou2017}. With the increasing use of transformers in NLP, recent research studies in the field of speech recognition mainly use transformers in their architecture. For multilingual speech recognition, \citep{Zhou2018} proposed a sequence-to-sequence attention-based model with a single Transformer that uses sub-word without using any pronunciation lexicon for their model.

\subsection{Sentiment Analysis}
Sentiment analysis aims to identify and extract information, such as feelings, attitudes, emotions, and opinions, from a piece of text. Many businesses use this service to improve the quality of their product by analyzing the comments of their customers. One of the main challenges for this task is reaching acceptable performance for those languages with lower resources. To tackle this challenge, \citep{Can2018} trained a model on a high resource language (English) and reused it for four other languages (Russian, Spanish, Turkish, and Dutch) with more limited data in the field of sentiment analysis. A novel deep learning method has been proposed in \citep{Nankani2020}. This work discusses the significant challenges involved in multilingual sentiment analysis and other methods for estimating sentiment polarity in multilingual for overcoming the problem of excessive dependence on external resources introduced \citep{Lin2014}.  
Also, in \citep{Kanclerz2020} a novel technique for the use of language-agnostic sentence representations to adapt the model trained on texts in Polish (as a low-resource language) is presented to recognize polarity in texts in other (high-resource) languages.

\subsection{Intent detection and Slot filling}
Intent Detection is the task of detecting users' messages and assigning suitable labels using chat-bots and intelligent systems, and slot filling tries to extract the values of certain types of attributes \citep{Huang2017}. Studies show that there is a strong relationship between these two terms, which leads to achieving state-of-the-art performance \citep{Weld2021}. Models in this field usually use joint deep learning architectures in attention-based recurrent frameworks. In \citep{Castellucci2019}, \citep{8907842} a "recurrence-less" model using BERT-Join was proposed that showed strong performances for these tasks. Also, they reached a similar performance for the Italian language without changing the model.

\subsection{Dependency Parsing}
Dependency parsing is another big challenge, especially in the field of multilingual NLP. \citep{Wang2020} used the BERT transformation approach to generate cross-lingual contextualized word embeddings. This linear transformation learned from contextual word alignments is trained in different languages and showed effectiveness on zero-shot cross-lingual transfer parsing and proved that this method outperforms static embeddings.

\subsection{NER}
NER is the task of extracting entities in the text and categorizing them into predetermined categories. Recent self-attention modes presented a state-of-the-art performance in this task, especially for inputs consisting of several sentences. This property became more important when it comes to analyzing data in several languages. \citep{Luoma2020} with using BERT in five languages, explores the use of cross-sentence information for NER and shows outperforming NER on all of the tested languages and models.
For languages with little or no labeled data, \citep{Wu2020a} proposed a teacher-student learning method for addressing this problem in both single-source and multi-source cross-lingual NER.

For evaluating different architectures for the task of name transliteration in a many-to-one multilingual paradigm such as LSTM, biLSTM, GRU, and Transformer, \citep{Moran2020} shows improving accuracy in transformer architecture for both encoder and decoder.

\subsection{Question Answering }
Question Answering (QA) is the task of building an automatic system to answer questions posed by humans in a natural language \citep{Ranjan2021}. This task is gaining lots of attention, especially in the field of multilingualism, but it is very challenging too. Different approaches for constructing meaning are used in various languages. For example, for the plural form of words in English, we usually use 's' in the end, but in Arabic, the plural form of words is not just adding postfix to words and sometimes the whole structure of word changes. Or some other languages like Japanese don't use space between words \citep{Clark2020}.

\section{Challenges and Outlook}
\label{sec:challenges}
This section provides some of the challenges in the domain of multi-lingual tasks and a set of ideas to be considered as future direction of this research line.

\subsection{Existing Challenges}
We identified three group of challenges in the domain of using transfer learning for multilingual tasks including challenges on (i) Modeling, (ii) practical aspects and (iii) applications. Next we provide details on each group of challenges.   
\subsubsection{On Modeling aspect}
challenges of pre-trained models due to the complexity of natural language processing can be grouped as follow:
\begin{itemize}
    \item Various objective tasks that evaluate different features of models. A challenging objective task can help in the manner of creating more general models. However, these tasks should be self-supervised because many captured corpora do not have tagged data.
    \item Due to the increasing use and research on multilingual and cross-lingual models, their vulnerability and reliability have become very important. In Section \ref{mlperformamce}, we reviewed some researches in this area and noted the less studied multilingual models. Nowadays, most of the researches in this category, conducted on mBert.
\end{itemize}

\subsubsection{On Practical aspects}
Research studies on following problems are affected by the high cost of pre-training models:
	\begin{itemize}
		\item General purpose models can learn the fundamental understanding of languages. However, usually need more profound architecture, larger corpora, and Innovative  pre-training tasks.
		\item Recent studies have confirmed the performance of Transformers in pre-trained models. Nevertheless, the computational resource requirement of these models limits their application. Therefore, model architecture improvement needs more attention in the research area. Moreover, architecture improvements could lead to a better contextual understanding of the language model, as it could deal with a more extended sequence and recognize context. \citep{Zoph2016} 
		\item Achieve maximum performance of current models: Most existing models can improve performance with increasing model depth, for example, with a more comprehensive input corpus or train steps. 
	\end{itemize}

\subsubsection{On Application}

	\begin{itemize}
	\item In terms of multilingual tasks, many task do not have enough data resources to gain significant performance in a specific application.
	\item The next big challenge is to successfully execute NER, which is essential when training a machine to distinguish between simple vocabulary and named entities. In many instances, these entities are surrounded by dollar amounts, places, locations, numbers, time, etc., it is critical to make and express the connections between each of these elements, only then may a machine fully interpret a given text. 
	\item Another challenge to mention is extracting semantic meanings. Linguistic analysis of vocabulary terms might not be enough for a machine to correctly apply learned knowledge. To successfully apply learning, a machine must understand further, the semantics of every vocabulary term within the context of the document.
\end{itemize}

\subsection{Future Directions}

Based on this study and analysis of the existing efforts in the domain of multi-lingual tasks, the following can be considered as the future direction of this research domain:
\begin{itemize}
    \item \textit{Vertical extension:} performance improvement of current models by increasing the number of pre-train steps, total parameters number of models, and larger input corpora which further result in higher training costs. In this manner, the requirement of processing power is undeniable. Another suggestion is to analyze the relationship between these hyper parameters of the model and the resulting performance of each model. 
	
	\item \textit{Horizontal expansion:}performance of pre-trained language model are related to corpora and its variety, so we can suggest expanding current research studies with multilingual corpora pre-training and evaluation in multilingual downstream tasks. Same as Vertical extension changes that result in pre-training, a model usually requires remarkable processing units. 
	
	\item   One challenging study field would be pre-training tasks, especially in cross-lingual models. Any progress in this manner would result in a more comprehensive evaluation of models. 
	
	\item  Optimization of model architecture design or methods like training process is another deep research way. enhancements in this aspect can result models that could pre-train on massive multilingual corpora with current computing resources.

	\item  recently we can see tend to Specific purpose models,  for specific domain application like health advice, but there is a gap already in this direction for example in low-resource or real-time computing, need for a newly designed model with specific tasks pre-train objective can be seen.

	\item Robustness of Pre-train models also needs more attention. studies with this subject will provide good insight into the future of this model in the industry.
\end{itemize}

\section{Conclusion}
\label{sec:Conclusion}
This survey provides an comprehensive overview of the existing studies on leveraging transfer learning models to tackle the multi-lingual and cross-lingual tasks. In addition to the models, we also reviewed the main available datasets in the community and investigated different approaches in term of the architectures and applications to identify the existing research challenges in the domain and later we provide few potential future directions. 

\section{Acknowledgments}
Mostafa Salehi is supported by a grant from IPM, Iran (No. CS1400-4-268).

{\small
\Urlmuskip=0mu plus 1mu\relax
\bibliographystyle{plain}
\bibliography{ms}
}
\end{document}